\pdfoutput=1
\documentclass[10pt,twocolumn,letterpaper]{article}

\usepackage[final]{cvpr}      

\usepackage{graphicx}
\usepackage{amsmath}
\usepackage{amssymb}
\usepackage{booktabs}
\usepackage[pagebackref,breaklinks,colorlinks]{hyperref}
\usepackage[capitalize]{cleveref}
\crefname{section}{Sec.}{Secs.}
\Crefname{section}{Section}{Sections}
\Crefname{table}{Table}{Tables}
\crefname{table}{Tab.}{Tabs.}
\newcommand{\AN}{$\mathcal{A}$}
\newcommand{\AVN}{$\mathcal{AV}$}

\def\cvprPaperID{****} 
\def\confName{CVPR}
\def\confYear{2022}

\begin{document}

\title{Egocentric Deep Multi-Channel Audio-Visual Active Speaker Localization}

\author{Hao Jiang, Calvin Murdock, Vamsi Krishna Ithapu\\
Meta Reality Labs, Redmond, USA
}
\maketitle


\begin{abstract}
Augmented reality devices have the potential to enhance human perception 
and enable other assistive functionalities in complex conversational environments.
Effectively capturing the audio-visual context necessary for understanding these
social interactions first requires
detecting and localizing the voice activities of the
device wearer and the surrounding people.
These tasks are challenging due to their egocentric nature: 
the wearer's head motion may cause motion blur,  
surrounding people 
may appear in difficult viewing angles, and there may be occlusions, 
visual clutter, audio noise, and bad lighting.
Under these conditions, previous state-of-the-art 
active speaker detection methods do not give satisfactory results. 
Instead, we tackle the problem from a new setting using both video and multi-channel microphone array audio. 
We propose a novel end-to-end deep learning approach that is able to give robust voice activity detection
and localization results. 
In contrast to previous methods, 
our method localizes active speakers from all possible directions on the sphere, even outside the camera's field of view, 
while simultaneously detecting the device wearer's own voice activity. 
Our experiments show that the proposed method gives superior results, can run in real time, and 
is robust against noise and clutter.

\end{abstract}	


\vspace{-10pt}
\section{Introduction}
\vspace{-5pt}

\begin{figure}[tb]
\centering
\includegraphics[width=0.325\linewidth]{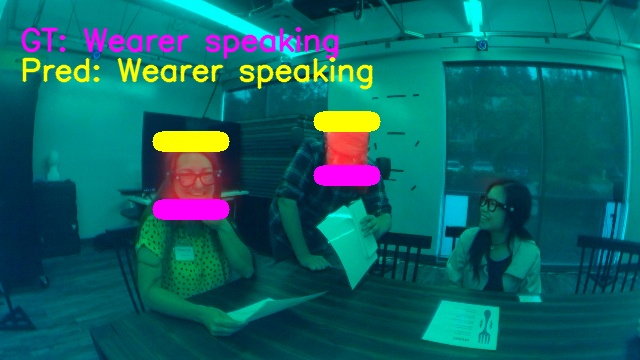}\hspace{\fill}
\includegraphics[width=0.325\linewidth]{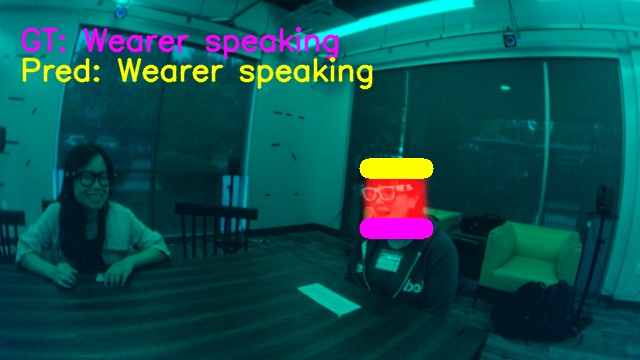}\hspace{\fill}
\includegraphics[width=0.325\linewidth]{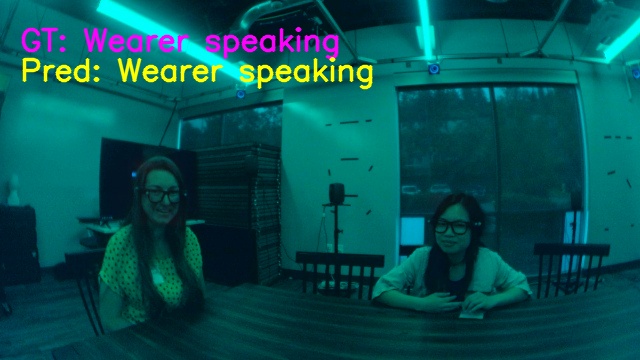}%
\linebreak
\includegraphics[width=0.325\linewidth]{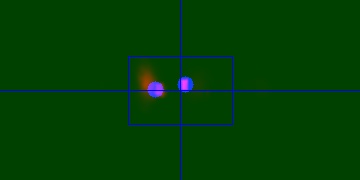}\hspace{\fill}
\includegraphics[width=0.325\linewidth]{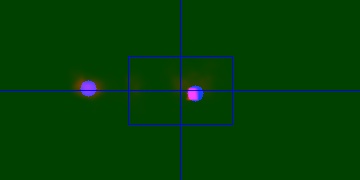}\hspace{\fill}
\includegraphics[width=0.325\linewidth]{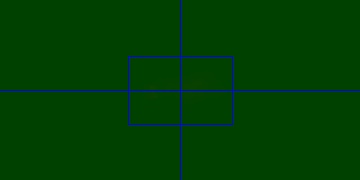}%
\vspace{0.2em}

\includegraphics[width=0.325\linewidth]{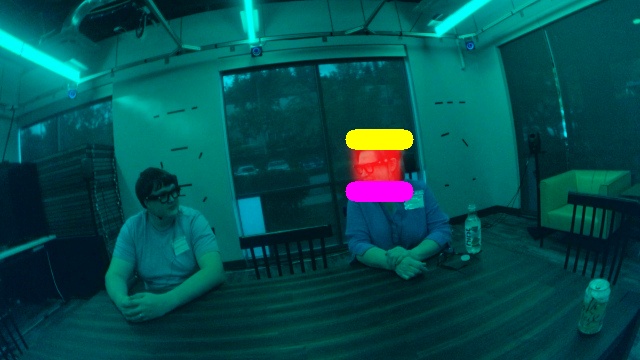}\hspace{\fill}
\includegraphics[width=0.325\linewidth]{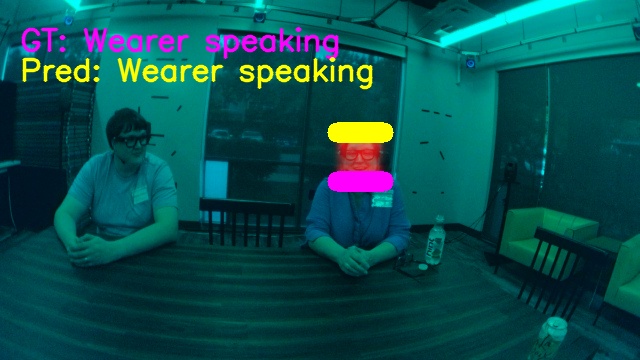}\hspace{\fill}
\includegraphics[width=0.325\linewidth]{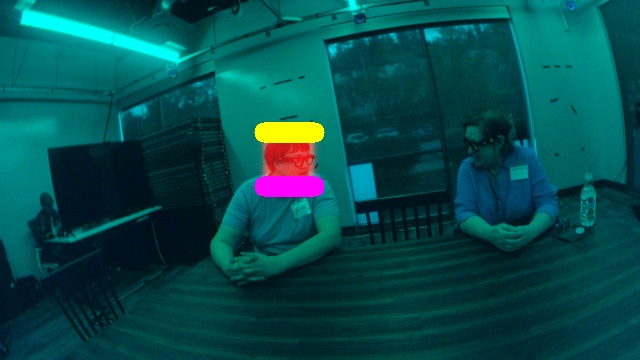}%
\linebreak
\includegraphics[width=0.325\linewidth]{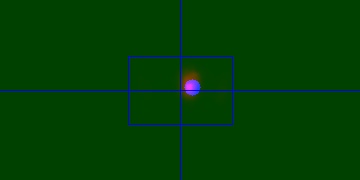}\hspace{\fill}
\includegraphics[width=0.325\linewidth]{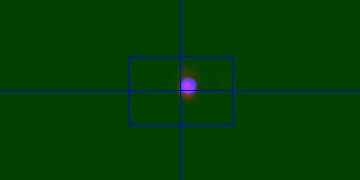}\hspace{\fill}
\includegraphics[width=0.325\linewidth]{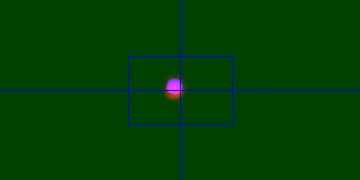}%
\vspace{-5pt}
\caption{Our novel multi-channel audio-visual deep network localizes active speakers from any direction on the sphere. 
	In this illustration, predicted active speaker probability heat maps are shown in the red channel of the images (rows 1,3) 
	alongside the full 360$^{\circ}$ voice map (rows 2,4) where the camera's limited field of view is indicated in the central blue rectangle. 
	This 360$\times$180 voice map (rows 2,4) is a cylindrical 2D projection of the sphere
	where each pixel corresponds to a direction in the device wearer's local 3D coordinate system. 
	The ground truth is shown as the purple bar under the talking head's 
	lower edge and the blue dots in the 360$^{\circ}$ map. The yellow bar at the upper edge of a head box shows
	our prediction that the person is talking. Our method also predicts whether the device wearer speaks.
	}
\vspace{-10pt}	
\label{fig:teaser}	
\end{figure}

Understanding conversational context and dynamics from an egocentric perspective is vital for creating realistic and useful augmented reality (AR) experiences. 
These attributes characterize the interactions of multiple speakers in a given scene with the AR device wearer (i.e., {\it ego}). 
An example such device may consist of glasses with outward looking cameras and microphones so that audio-visual data is captured from the wearer's point of view. 
Modeling these attributes involves not only 
detecting and tracking people within a scene, but also localizing the voice activity within a conversation.
In this work, we focus on the task of active speaker localization (ASL) with the  
goal of detecting the spatio-temporal location of all active speakers both within and outside the camera's field of view (FOV). 
Closely related to the problem of active speaker detection (ASD), ASL involves estimating 
the relative direction of arrival of speech from an egocentric perspective. 
In this paper, active speakers typically correspond to the people who are speaking and `driving' the conversations.
The elements of our proposed egocentric ASL problem are illustrated in Fig.~\ref{fig:teaser}.

A good ASL system needs to account for the changing orientations of speakers from egocentric point of view
and be robust to speakers moving in and out of the visual field of view.
In particular, natural conversations entail significant overlap between different speakers' voice activity 
and involve one or more speakers interrupting each other --- a classical attribute in conversational ecology called turn-taking.    
Such a system should also ideally be agnostic to the number of microphone channels, 
thereby allowing for generalization to different AR devices with varying numbers of audio and/or visual channels.  
Note that the device wearer may also be an active speaker during the conversation whose voice is naturally amplified 
due to their closeness to the device microphones. 
An ASL system must account for this {\it false} amplification that may nullify competing active speakers in the scene.
In this work, we propose a real-time audio-visual ASL system that addresses these aspects to 
effectively localize active speakers potentially outside of the visual FOV by leveraging audio recorded from a device-mounted
microphone array. 

We propose a new end-to-end deep neural network trained to tackle this problem. 
Our network is partitioned into two branches: an audio network and an audio-visual network. 
The audio network builds useful representations for constructing a low-resolution sound source localization map with a full 360$^{\circ}$ FOV 
by utilizing spatio-temporal correlations across different channels. 
The audio-visual network then combines the extracted audio features with the corresponding video frames, 
resulting in a higher resolution activity map for the camera's FOV. 
Visual cues such as the person's mouth movement, facial expressions, and body pose are extracted here 
and combined with audio features for computing a joint representation. 
The final 360$^{\circ}$ active speaker map is a combination of the low-resolution audio-only map and the high-resolution audio-visual map.
In addition, the device wearer voice detector shares the features from the audio network, 
and our model estimates the relative 3D orientations of the speakers in the scene from egocentric perspective. 
The proposed network is also aimed at real-time applications in the immersion-driven domain of AR, 
enabling systems for the spatialization and localization of audio-visual activity in world-locked frame of reference.
Lastly, the lack of reliable multi-channel conversational datasets is another limiting factor for building in-the-wild ASL systems. 
To that end, we build and evaluate our approach using a very recent egocentric conversations dataset called EasyCom \cite{easycom}. 

Our contributions are: 
\begin{enumerate}
	\item We tackle the new problem of ASL using multi-channel audio and video from egocentric perspective. In this new problem, we localize 
	      all the active speakers in the scene including the device wearer. 
	\item We propose a real-time low-latency egocentric audio-visual active speaker localization system with a 360$^{\circ}$ field of view. 
                  Our novel deep multi-channel audio-visual network learns from different audio features 
		  and can accommodate different numbers of audio channels without structure changes.
	\item We evaluate our method on the EasyCom dataset and demonstrate significantly 
	improved results in comparison to previous audio-visual ASL and ASD approaches.  
\end{enumerate}


\subsection{Related Work} \label{sec:related}
Single and multi-channel sound source detection and localization problems have classically been studied 
by speech and audio signal processing communities \cite{audioprocess, doa, soundloc1}. 
Most of these works are based on source separation and voice activity detection, 
and they mainly assume that there is one speaker in the audio stream who dominates the others (i.e., a high signal-to-noise ratio). 
The primary characteristic of these methods is to build auto-correlation and cross-correlation functions across different channels
to account for timing and level differences caused by microphone placement.  
However, these approaches are sensitive to room acoustics and noisy backgrounds and may be unreliable when multiple sources are present. 
More recently, machine learning has been used for direction of arrival estimation with some success \cite{deep-sound-loc1, deep-sound-loc2, deepdoasurvey}. 
Although these methods improve upon the traditional approaches, 
the lack of visual information limits the efficacy of these systems in real-word settings. 
Furthermore, most multi-channel approaches assume fixed, stationary microphone arrays, 
which may lead to poor performance with moving arrays in egocentric settings. 

The computer vision community has seen a surge in audio-visual learning research, 
in particular due to datasets like the AVA Speech and Activity corpus \cite{avadataset}, Voxconverse \cite{voxconverse}, and Voxceleb \cite{voxceleb2}. 
These approaches are driven by building correspondences between audio and visual modalities, thereby resulting in 
robust joint representations that improve upon their audio-only or image-only counterparts. 
For action and activity recognition, several studies have shown evidence that audio disambiguates certain visually ambiguous cues 
\cite{kazakos2019epic,audiovisual-slowfast}. 
Audio-visual models have been explored for speech recognition \cite{avspeech}, 
sound source detection \cite{avloc1, avloc2, avloc3}, multiple source separation \cite{avsep2, avsep4, avsep5, avsep6}, 
localization of sounds in a 2D image \cite{360sound, avsep1}, 3D scene navigation guided by audio \cite{avnavi}, and others. 

A bulk of the audio-visual learning models follow a simple recipe: 
audio inputs are often converted to spectrogram images which are then jointly processed with video frames.
In addition to traditional network architectures, 
transformer networks have also been proposed for single-channel active speaker detection \cite{talknet}.
More recently, turn-taking has also been studied as a means to improve detection performance \cite{iccv21b}.
A related problem is that of speech separation, which singles out
a speaker's voice by using both audio and cropped facial images \cite{avsep2, avsep5, avsep6}. 
The voice energy of the enhanced speech can then be used to detect active speakers.
Although extensively studied, single-channel speaker detection from an egocentric perspective is still a challenging problem. 
This is mainly because of substantial device motion, 
occlusions, reduced visibility of speakers' faces, and noise induced by overlapping and interrupting speakers. 
Most current methods also induce significant latency in detection, which would be ineffective for enabling real-time AR experiences. 

Single-channel audio-visual localization in exocentric 
settings has received much attention lately 
\cite{av1, avloc1, avloc2, avloc3, iccv21a}.
These methods either utilize audio-visual joint embeddings similar to those in active speaker detection, 
or they train audio-visual joint classification modules as the backbone for modality fusion. 
In addition, due to the lack of multiple channels, localization is restricted to the image frame 
in a manner similar to traditional visual object localization.
The most recent related work is from \cite{wangaaai}, where the authors propose an audio-visual model 
that can process binaural (two-channel) audio for sound source localization. 
However, the system cannot be extended to multi-channel settings, and is restricted to localizing targets within the visual field of view. 


\section{Egocentric Active Speaker Localization} \label{sec:framework}

\paragraph{Problem Setup:}
Given multi-channel audio-visual data captured using AR glasses with a microphone array and RGB camera, 
we define the egocentric ASL problem as the detection and spatio-temporal localization of all the active speakers in the scene including the voice activity of the device wearer. 
Let $\mathbf{A}_i$ with ($i=1..N$) denote the audio signals captured via $N$-channel microphone array and $\bf{I}$ denote the video from the RGB camera.
The audio signals are normalized to the range [-1,1] based on the maximum bit length of audio samples.
At each time instant $t$, given a segment of audio ${\mathbf{A}^t_i}$ and the corresponding video frame $\mathbf{I}^t$,
we estimate two outputs: a heat map $\mathbf{V}^t_{\alpha,\beta}$ of activity in the scene and the device wearer activity $\bf{W}$. 
$\mathbf{V}^t_{\alpha,\beta}$ is a 2D matrix where each element gives the probability of a sound source being present at 
particular relative angles $(\alpha,\beta)$ at the time instant $t$, 
where $\alpha \in [-180,180]$ and $\beta \in [-90,90]$ correspond to azimuthal (horizontal) elevation (vertical) respectively. 
Although we focus on human voice in this work, the proposed framework is applicable to any sounds of interest. 

\subsection{Overview} \label{sec:overview}

\begin{figure*}[tb]
\centering
\includegraphics[width=0.85\linewidth]{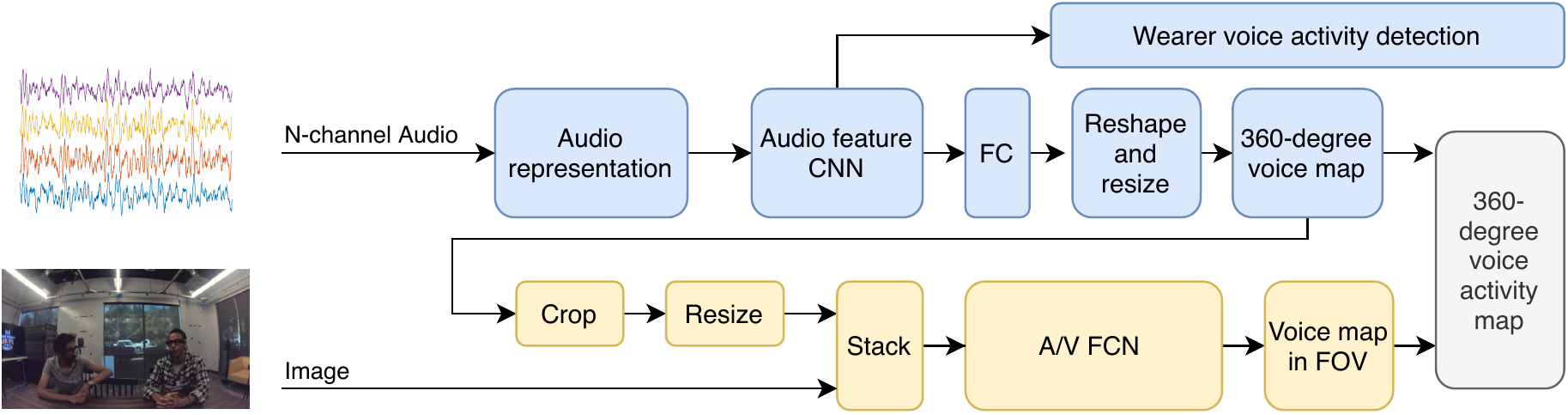}
\vspace{-5pt}
\caption{Egocentric multi-channel audio-visual localization. Our end-to-end deep network detects a 360$^{\circ}$ voice activity map
	and the wearer's voice activity at the same time.}
\vspace{-5pt}	
\label{fig:system}	
\end{figure*}

Fig.~\ref{fig:system} illustrates the proposed egocentric ASL framework. 
Our method is an end-to-end deep learning model which takes the raw audio and video as input and estimates the 
active speaker activity heat map ($\bf{V}$) and wearer's voice activity ($\bf{W}$) directly. 
The framework has two networks: an audio network cascade (\AN) and an audio-visual network cascade (\AVN). 
\AN $ $ converts raw multi-channel audio and compacts a 2D representation aligned to each video frame, 
which is then used to extract relevant features using a convolutional neural network to estimate a direction of arrival estimate for the sources in the scene. 
\AVN $ $ then utilizes the outputs from \AN $ $ and incorporates visual information using another network. 
The resulting outputs from both \AN $ $ and \AVN $ $ are then combined to compute the scene and wearer's activity ($\bf{V}$ and $\bf{W}$).


\subsection{Audio Network} \label{sec:audio-only}

\vspace{-5pt}
\subsubsection{Audio Representation}
\vspace{-5pt}
In this paper, we consider three audio representations and design our deep network so that
it can take these different representations together with video as input in the same fashion.
Our experiments show these audio representations are stronger than the raw audio. These different audio representations
have different properties that are suitable for different use cases.

Our first audio representation is adapted from the complex spectrogram representation \cite{wangaaai}.
For audio with sampling rate of $48kHz$ and video frame rate at $20Hz$, we compute the short-time Fourier
transform (STFT) extract 
100 discrete Fourier transforms (DFTs) of length 200 to align with each video frame. 
The real and imaginary parts of the DFTs from all the channels are stacked together
along the depth axis
to form the multi-channel 2D tensor. 

\begin{figure}[tb]
\centering
\includegraphics[height=1.3cm]{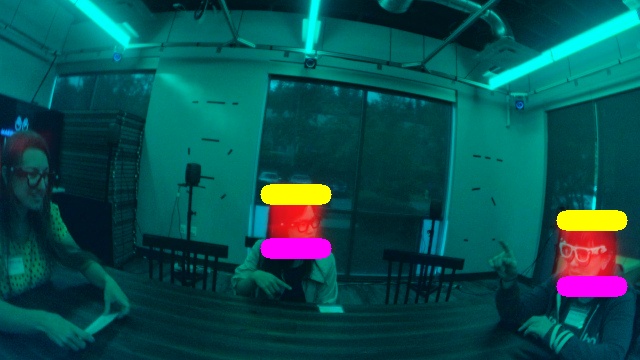}%
\includegraphics[height=1.3cm]{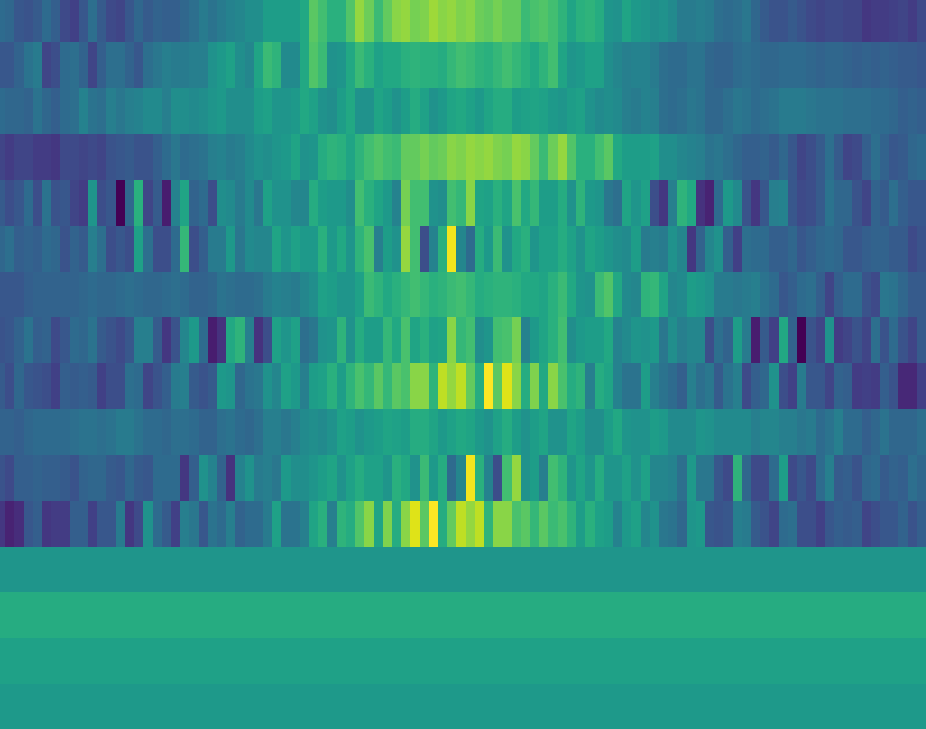}\hspace{0.02em}
\includegraphics[height=1.3cm]{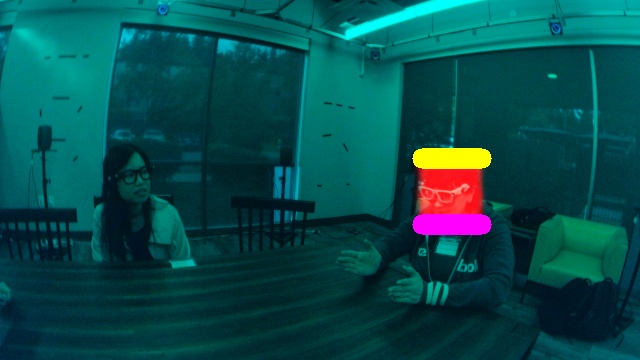}%
\includegraphics[height=1.3cm]{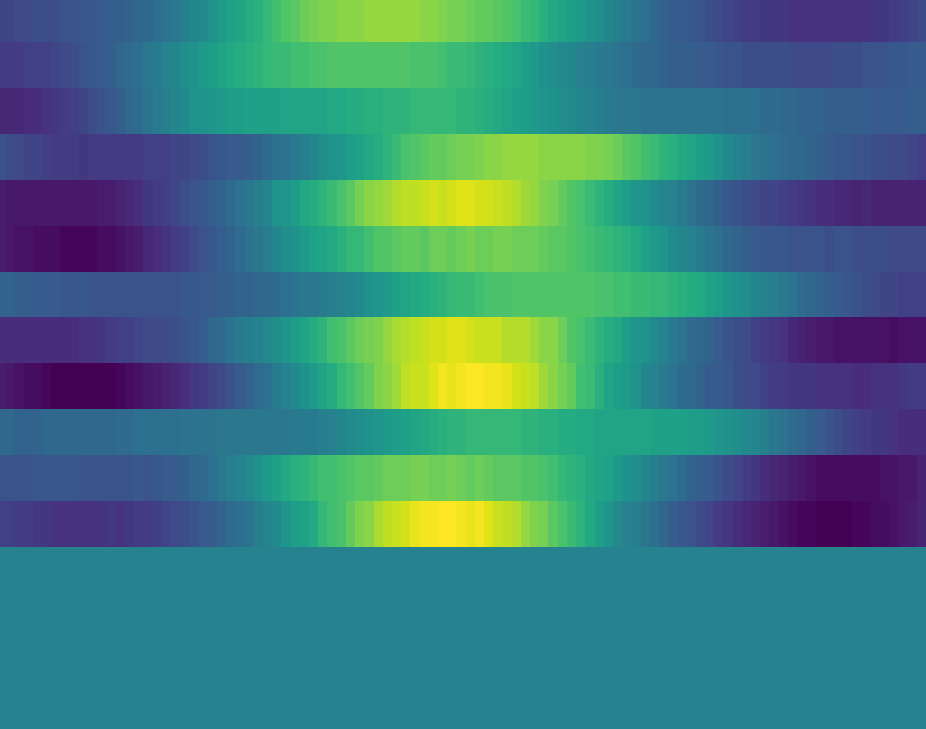}%

\vspace{0.15em}

\includegraphics[height=1.3cm]{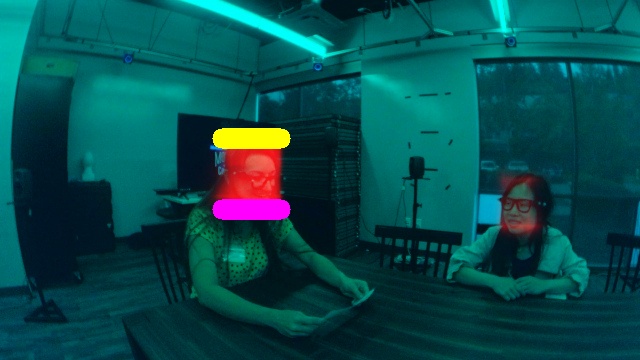}%
\includegraphics[height=1.3cm]{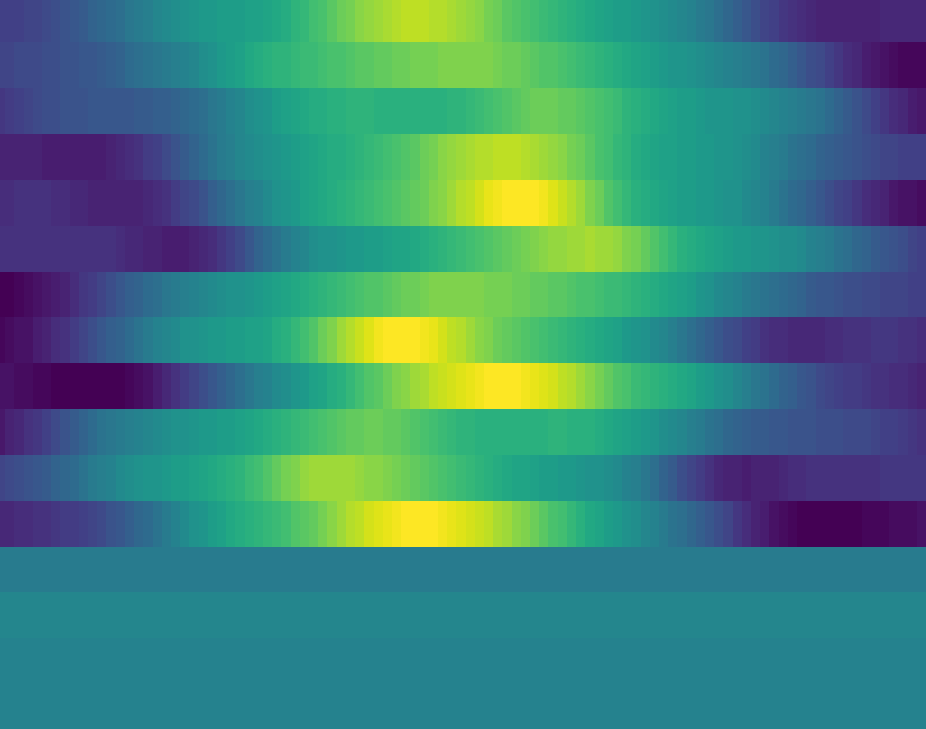}\hspace{0.02em}
\includegraphics[height=1.3cm]{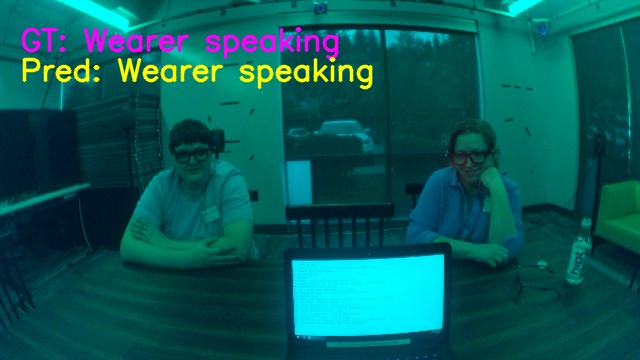}%
\includegraphics[height=1.3cm]{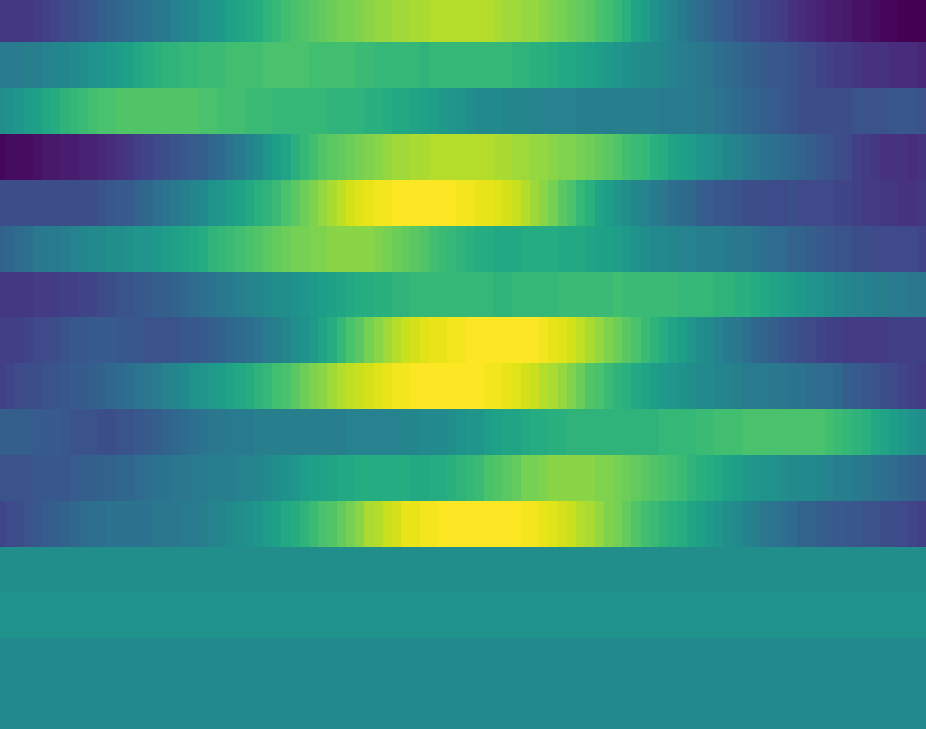}%
\vspace{-5pt}
\caption{Odd columns: video frames overlaid with voice activity labels. 
Even columns: vertical stack of the
	audio cross correlation and energy feature maps.}
\vspace{-20pt}	
\label{fig:feature}
\end{figure}

Apart from complex spectrogram,  
we further propose a 2D audio representation that captures the cross correlation between all pairs of the 
audio channels. 
Unlike spectrograms, this representation is mostly speaker invariant.
In more details, assuming the audio sample $n$ matches the time stamp of video frame at time $t$, the cross
correlation $C_{p,q}(n,m)$ between channel $p$ and $q$ is
\vspace{-3pt}
\small
\[
	C_{p,q}(n, m) = \frac{\sum_{k=0}^K [A_p(n - k) A_q(n - k + m)]}{\sqrt{\sum_{k=0}^K A_p(n - k)^2)}\sqrt{\sum_{k=0}^K A_q(n - k + m)^2)}}, 
	\vspace{-3pt}
\]
\normalsize
where $m=[-L,L]$, and $K$ and $L$ are two parameters. 
In our experiments, audio signals have sampling rate $48kHz$, $K=1200 $ and $L=50$. 
In a discrete format, $C_{p,q}(n,m)$ is a vector of length $2L+1$ at each time $n$ 
that characterizes not only the time shift of different audio channels due to the different path of the sound
transmission, but also other fine-grained couplings between different audio channels.
Using this $C$, we construct a 2D audio representation at each time $n$, 
which is a stack of all the vectors $C_{p,q}(n,m)$ for each $(p,q)$ pair.

The short-time energy of audio is a feature that is invariant to sound sources 
and easy to compute.
Therefore, we also include a separate measure of the energies from each audio channel, 
\small $E_p(n) = (\sum_{k=0}^K A_p(n - k)^2)^{0.5}$. \normalsize
Using the $E$, we stack $\bf{e}_p(n)$ for each $p$, where $\bf{e}_p(n)$ is 
a vector that duplicates the $E_p(n)$ by $2L+1$ times to form a 2D energy map.
These features can also be combined to form richer representations. 
Fig.~\ref{fig:feature} illustrates how the combined cross correlation and energy feature correspond to 
the audio events in videos. 
The cross-correlation, energy and the combined 2D feature are further
resized. In this paper, the width and height are resized to 128. 

\vspace{-10pt}
\subsubsection{Audio Activity Network}
\vspace{-5pt}

\begin{figure}[tb]
\centering
\includegraphics[width=0.8 \linewidth]{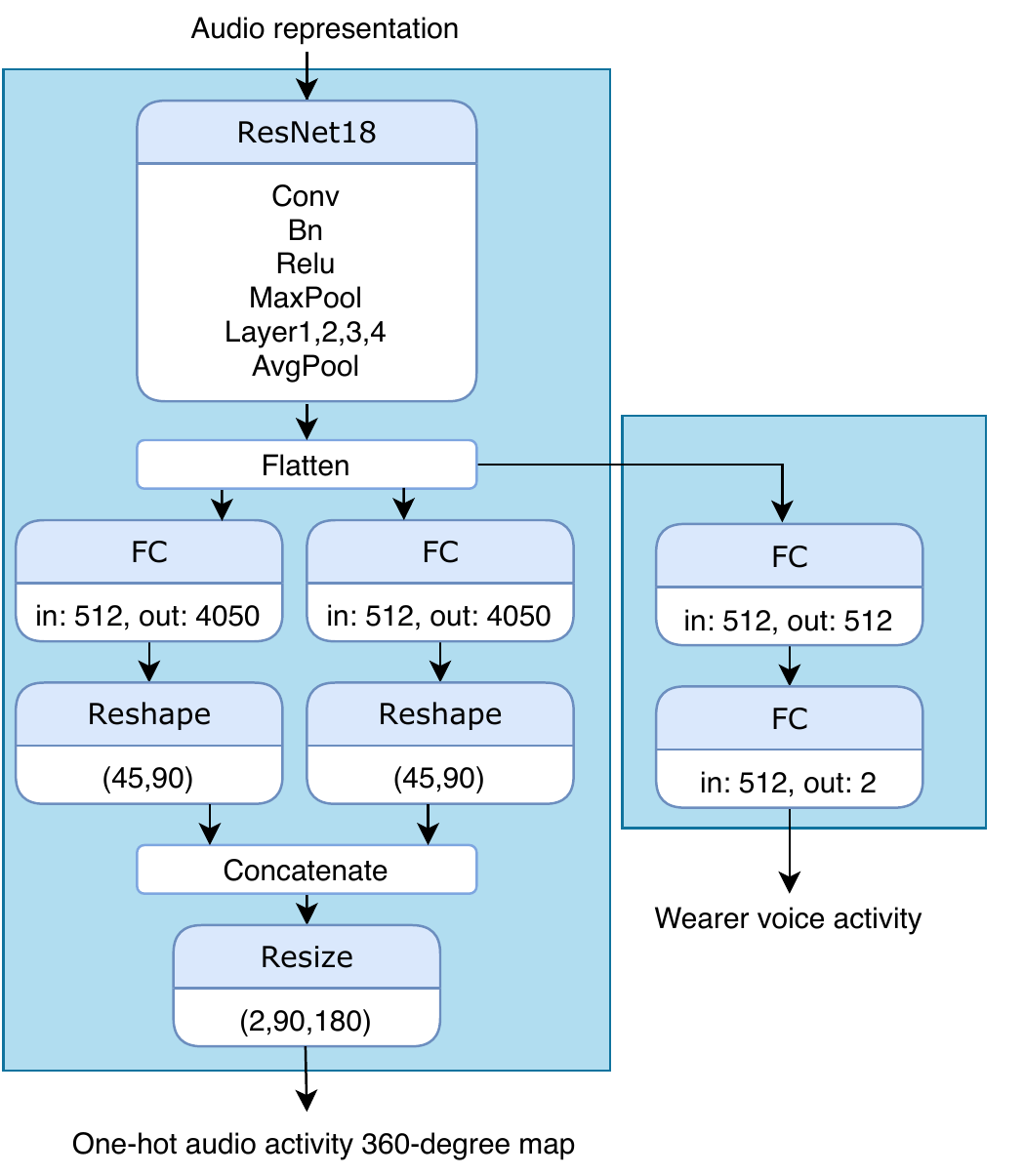}
\vspace{-5pt}
\caption{The audio activity network.}
\vspace{-10pt}
\label{fig:audio-doa}	
\end{figure}

The audio activity network predicts a rough 360$^{\circ}$ audio activity map and the voice activity of the device wearer. 
Its structure is shown in Fig.~\ref{fig:audio-doa}.
The feature extraction network is adapted from the first several layers of a ResNet18 network whose coefficients
are pre-trained on ImageNet. The first convolutional layer is modified to match the channel number of 
different audio representations.
The feature extraction network maps the audio 2D representation to a compact feature, which quantifies 
the spatial and voice characteristics of audio signals in the scene.
The extracted features are flattened and passed to two fully connected layers, which are further
reshaped to two $90\times45$ maps. The two maps are stacked and resized to a $180\times90$ one-hot representation
half the size of the full 360$^{\circ}$ audio activity map. This network thus predicts the voice activity probability 
from each direction on the sphere with an angular resolution of 2$^{\circ}$. 

One key design here is to generate the one-hot representation of the heat map and train using cross-entropy loss. 
This gives more stable results than  
directly regressing a single heat map of the audio activity using L1 or L2 losses.

The audio activity map is also used to simultaneously estimate the wearer's voice activity.
Due to the spatial position of the wearer's mouth relative to the microphones and the loudness
of the wearer's voice, the 2D feature representation learned by the audio localization network also 
provides useful information for detecting whether the device wearer is speaking.
To accomplish this, the audio feature extraction is shared with the 360$^{\circ}$ audio map prediction,
and wearer voice activity detection is performed by a separate head that consists of two fully-connected layers
trained to predict probability with a cross-entropy loss.

\subsection{Audio-Visual Network} \label{sec:audio-visual}

With only multi-channel audio available for speaker localization, the spatial resolution is low. This is due to the
inherent physics of sound propagation and the limitations of compact microphone arrays. 
We therefore also take advantage of video frames to further improve the estimation result. 
Images not only increase spatial resolution, but also provide extra clues related to voice activity, such as 
mouth movement, facial expression, and hand gestures.

\begin{figure}[tb]
\centering
\includegraphics[width=0.4 \textwidth]{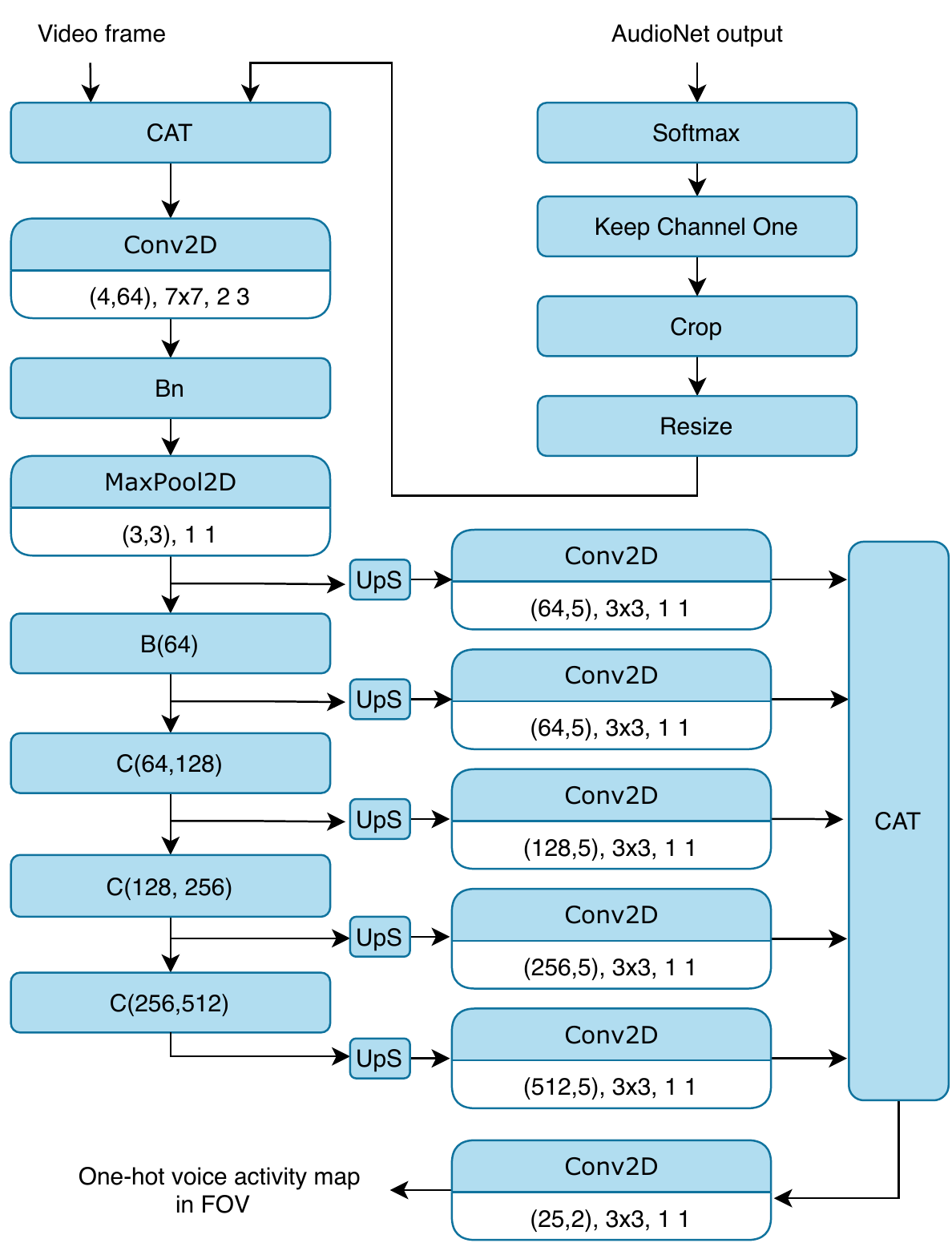}
\vspace{-5pt}
	\caption{Audio-visual network. The blocks $B(p)$ and $C(p,q)$ are defined in Fig.~\ref{fig:blocks}.
	For 2D convolution layers, the parameters are input channel number, output channel number, 
	convolution kernel size, stride and padding. For maxpool layer, the parameters are pooling kernel size,
	stride and padding.
	} 
\vspace{-10pt}	
\label{fig:audio-video-doa}	
\end{figure}

\begin{figure}[tb]
\centering
\includegraphics[width=0.5 \textwidth]{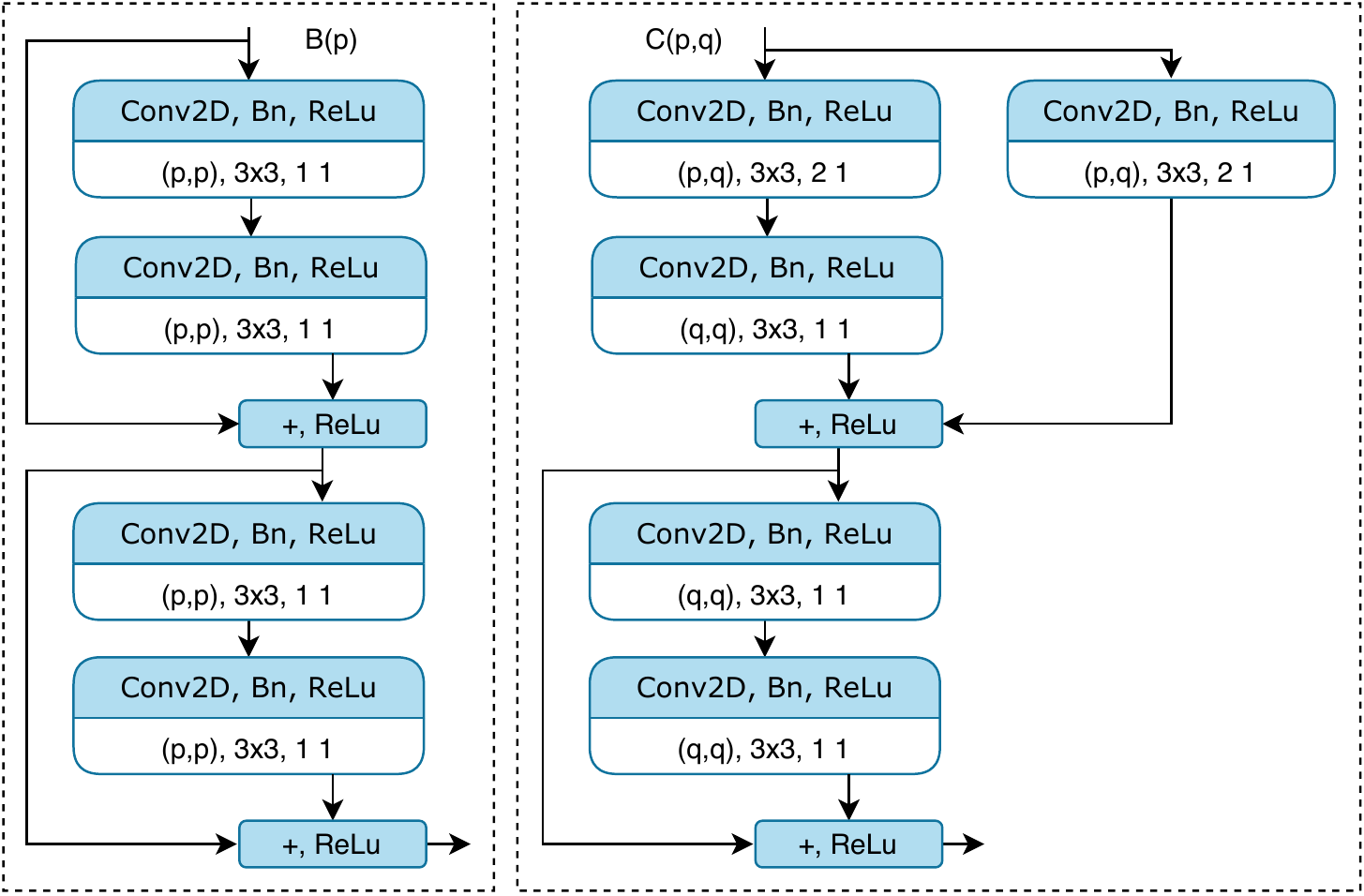}
\vspace{-20pt}
\caption{Residual blocks in the audio-visual network.}
\vspace{-20pt}
\label{fig:blocks}	
\end{figure}

In this paper, we propose a different approach to fusing audio and visual information
from previous audio-visual methods: we directly stack the
video frames with the estimated voice activity map from the audio network. Since the rough 360$^{\circ}$ voice map from the audio network 
is defined on the unit sphere and the grids are horizontal and vertical angles, we need a procedure to align the 
audio map to the corresponding video frames. Even though we can map each grid in the voice map to the image, 
we find a simpler cropping and scaling method is sufficient due to the low resolution of the 
audio map. 
More specifically, we crop the region from the audio map within the 
horizontal and vertical angles corresponding to the four corners of the image. 
The scaling procedure then upsamples the region so that the audio map in the FOV is aligned with the 
input video. These operations are integrated in the audio-visual network.
As shown in Fig.~\ref{fig:audio-video-doa}, the fused audio map and the corresponding color video frame form a tensor with depth of 4,
which is sent to a fully-convolutional network to estimate the refined voice activity map in the camera's field of view.
In this paper, the video resolution is $640\times360$. 

With such a design, 
if the faces are visible, the audio-visual network is able to take advantage of image features such as the appearance 
of the mouth and facial expression to localize audio activity. Due to its wide effective receptive field, 
the proposed network can also learn to extract other visual features such as body pose. 
Unlike previous methods, if the faces are not visible, our proposed method can still function because
the audio activity map gives the locations of the potential speakers in the scene.

We combine the rough 360$^{\circ}$ heat map and the more detailed heat map in the FOV. In this paper, we simply pad 
the refined heat map with zeros outside the camera's FOV and add it to the 360$^{\circ}$ heat map to generate the final estimation.

\subsection{Model Training} \label{sec:training}

We train the network in two stages. 
In the first stage, we train the audio-only and audio-visual network together without the wearer's voice activity classification network. 
In the second stage, we fix the audio feature layer's weights and train  
the fully connected network to predict the wearer's voice activity. 

The 360$^{\circ}$ voice map and the voice map in the FOV are represented
differently in the ground truth. The 360$^{\circ}$  voice map is a 180$\times$90 2D map. If there is a speaker located at $(\alpha, \beta)$,
the ground truth voice map has a solid disk with radius 5 centered at the point. Such labeling is uniform for regions 
inside and outside of the field of view. In contrast, the voice map in the FOV has the same size as the video frames, and 
the active speaker in the field of view is labeled as a solid rectangle that covers the speaker's
head. Therefore inside the FOV, the detection also has an attribute of size which is related to the depth of the target. The
training losses are defined as follows.

The first stage loss function is
\[
	\mathcal{L}_a = \mathcal{H}(y_{a}, \hat{y}_{360}) +  
	              \mathcal{H}(y_{av}, \hat{y}_{fov}),
\]
and the second stage loss function is:
\[
	\mathcal{L}_b = \mathcal{H}(y_{w}, \hat{y}_{w}),
\]	
where $\mathcal{H}$ is the mean cross entropy, 
$y_{a}$ and $y_{av}$ are the one-hot output representations of the audio-only and audio-visual networks, 
$\hat{y}_{360}$ and $\hat{y}_{fov}$ are their corresponding ground truth audio maps, 
$y_w$ is the wearer speech activity prediction, and $\hat{y}_{w}$ is its ground truth label.
The training procedure generally converges quickly within 5 epochs.  


\section{Experiment Results}
\vspace{-5pt}

In this section, we evaluate the proposed method on real videos and compare it with different audio-visual approaches
for active speaker localization and wearer voice activity detection.
Since we consider a novel egocentric problem setting, 
there are no previous audio-visual methods that are directly applicable.
For comparison, we instead adapt our multi-channel audio and video inputs to other
approaches to similar problems.
We also experiment with variations of the proposed method to justify our design choice.

\subsection{Evaluation Dataset}
\vspace{-5pt}

We evaluate our method using the EasyCom \cite{easycom} dataset. EasyCom is a multi-channel audio-visual dataset that includes around 
6 hours of 
egocentric videos of conversations within a simulated noisy environment. The dataset is recorded using a microphone array and a RGB camera mounted on a pair of 
glasses. EasyCom is a challenging dataset with significant background noise, fast head motion, and motion blur. 
Participants may sit or walk around in the scene, and 
their faces and mouths are
not always visible due to occlusions.
 
There are six microphones used for recording: four fixed to the glasses and 
two placed within the ears of the participants.
In this paper, we use the
RGB egocentric video together with the multi-channel audio from the four fixed microphones in our experiments.
The dataset has 12 video sessions, each of which is about half an hour long. There may be 4, 5, or 6 participants 
including the camera wearer in each recording session.
We use sessions 1--3 for testing and the remaining 9 sessions for training.
For fair comparison, we report the best numbers 
for all competing models trained until convergence after a sufficiently large number of epochs.

\subsection{Methods in Evaluation}
\vspace{-5pt}

We compare the proposed method in different variations against other active speaker detection and localization methods.
The methods in the evaluation include:
\begin{itemize}
\vspace{-8pt}		
	\item  Our method and variations (
		\texttt{Ours AV([cor] + [eng] + [spec] + [box])}):
		Variations include different combinations of feature representations (cor: cross correlation, eng: energy, spec: spectrogram, 
		and box: head bounding boxes).
		In the variation that uses head bounding boxes, we set the background color outside of the detected head regions to black.
		We also evaluate the audio-only and video-only versions of our method 
		in which the video or audio branches are removed from our full model.

\vspace{-8pt}		
	\item \texttt{DOA+headbox}: A state-of-the-art signal processing method \cite{doa} for extracting spherical 
direction-of-arrival (DOA) energy maps from the 4 microphones on
the glasses combined with head detection bounding boxes for active speaker detection. 
This DOA estimation method was designed to achieve more robust results in 
highly reverberant settings compared to previous signal processing audio localization methods.
To detect active speakers in the field of view, we pool regions of the DOA map 
corresponding to directions within the detected head bounding boxes. 
If the DOA map accurately estimates sound arrival directions, 
then the 
head bounding boxes corresponding to active speakers will include higher energy values. 

\vspace{-8pt}
\item \texttt{DOA+image}: A deep neural network trained to localize active speakers
using both traditional signal processing DOA maps \cite{doa} 
and video frames as inputs. 
		The network is fully convolutional and has the same structure as the audio-visual network in our method.

\vspace{-8pt}	
\item \texttt{AV-rawaudio}: A deep neural network trained using multi-channel raw audio and video as the input.
		Aside from extracting audio features with 1D convolution layers, 
		the overall network architecture is the same as our approach.
\vspace{-8pt}
	\item Mouth region classifier (\texttt{MRC}): A visual-only method for classifying active speech 
	from cropped images of mouth regions extracted from a 68-point facial key point detector.
Such a scheme has been commonly used in active speaker detection. 
		A ResNet18 network is trained to classify the cropped mouth images. We test two cases: \texttt{MRC(AVA)}
		trained using the AVA active speaker detection dataset \cite{avadataset}, and \texttt{MRC(EasyCom)} only trained on EasyCom.

\vspace{-8pt}		
\item \texttt{TalkNet} \cite{talknet}: A transformer-based single-channel audio-visual active speaker detection method 
	that gave 
	state-of-the-art
results in the AVA active speaker detection challenge. 
		We use the method in two modes: \texttt{TalkNet(AVA)} trained on the AVA dataset and \texttt{TalkNet(EasyCom)} 
		trained on EasyCom.
\vspace{-8pt}
	\item \texttt{BinauralAVLocation} \cite{wangaaai}: A two-channel audio-visual method for sound source localization. 
		Since this method cannot be easily extended to settings with more than two asymmetric microphones, we use only the audio channels from the 
		two frontal microphones in our comparisons.

\end{itemize}

\subsection{Within-View Active Speaker Localization}

\begin{figure*}[tb]
\centering
\includegraphics[width=0.166\linewidth]{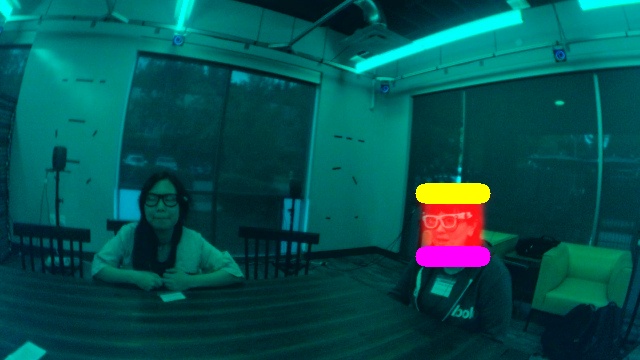}%
\includegraphics[width=0.166\linewidth]{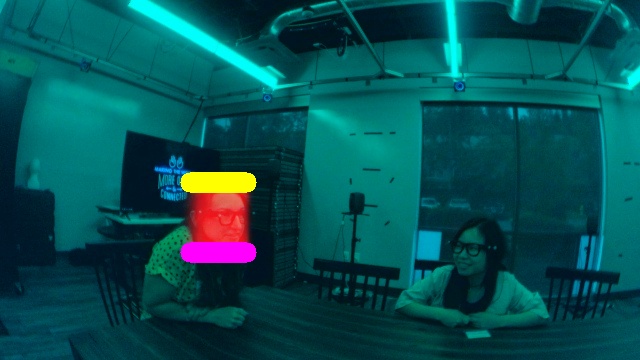}%
\includegraphics[width=0.166\linewidth]{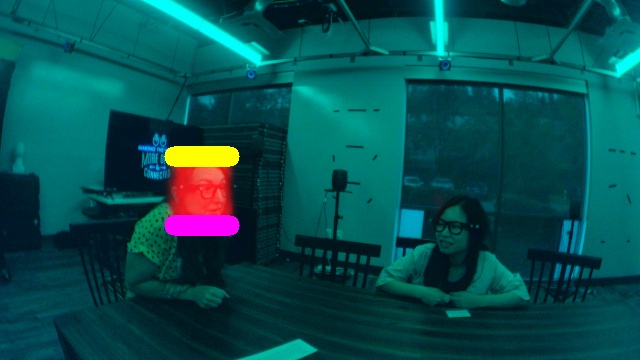}%
\includegraphics[width=0.166\linewidth]{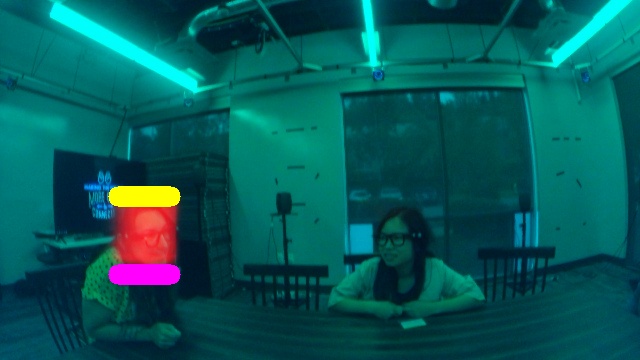}%
\includegraphics[width=0.166\linewidth]{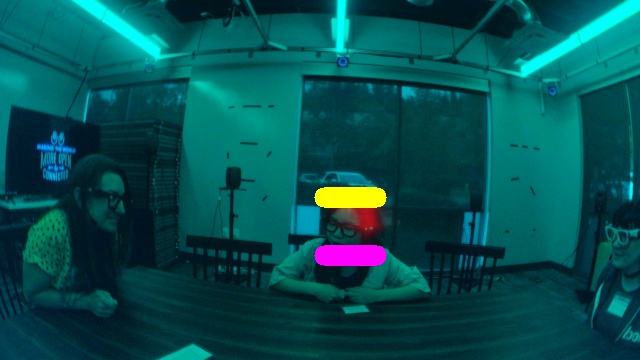}%
\includegraphics[width=0.166\linewidth]{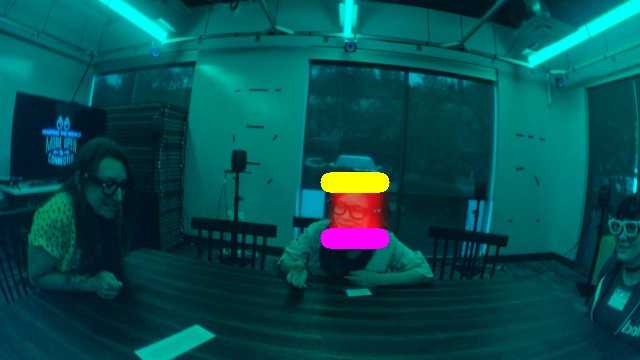}%
\linebreak
\includegraphics[width=0.166\linewidth]{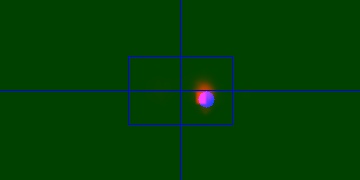}%
\includegraphics[width=0.166\linewidth]{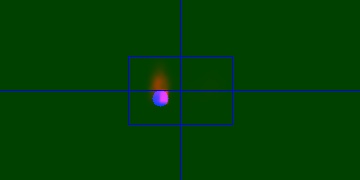}%
\includegraphics[width=0.166\linewidth]{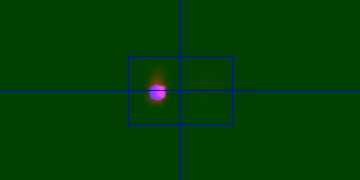}%
\includegraphics[width=0.166\linewidth]{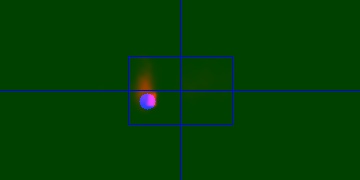}%
\includegraphics[width=0.166\linewidth]{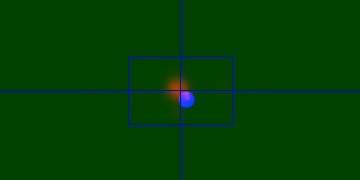}%
\includegraphics[width=0.166\linewidth]{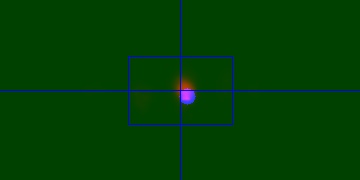}%
\linebreak
\includegraphics[width=0.166\linewidth]{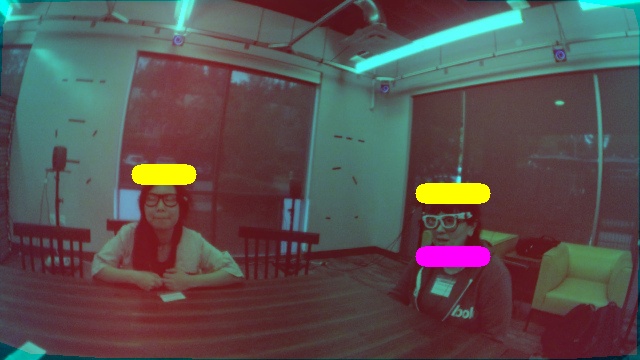}%
\includegraphics[width=0.166\linewidth]{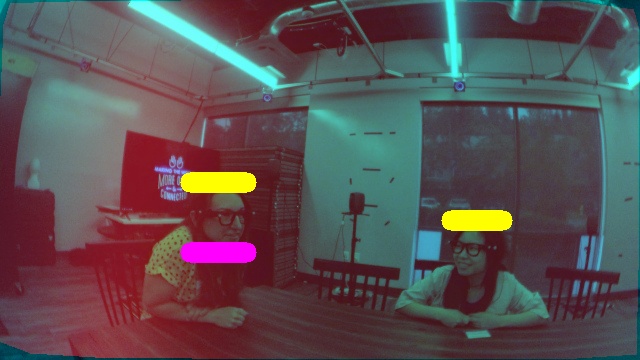}%
\includegraphics[width=0.166\linewidth]{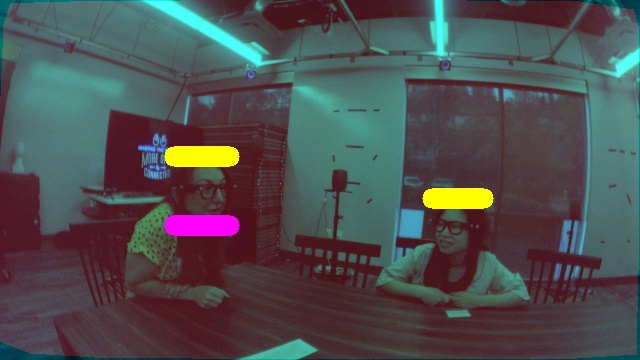}%
\includegraphics[width=0.166\linewidth]{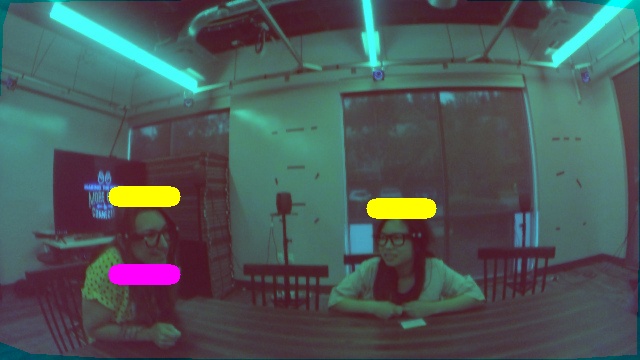}%
\includegraphics[width=0.166\linewidth]{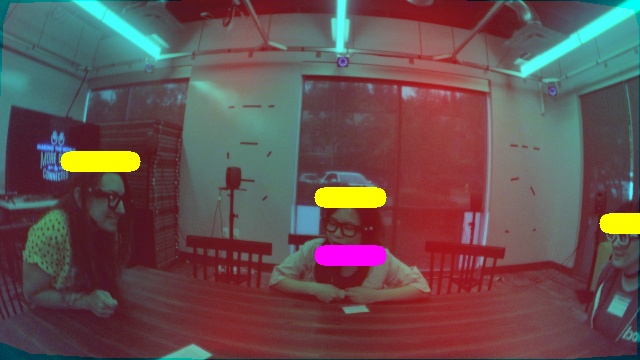}%
\includegraphics[width=0.166\linewidth]{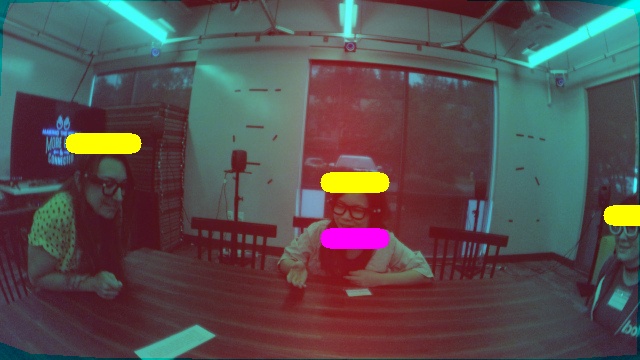}%
\linebreak
\includegraphics[width=0.166\linewidth]{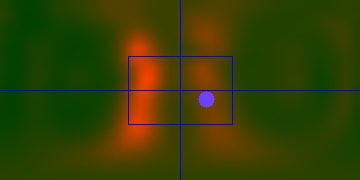}%
\includegraphics[width=0.166\linewidth]{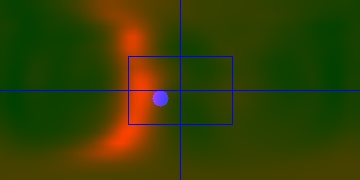}%
\includegraphics[width=0.166\linewidth]{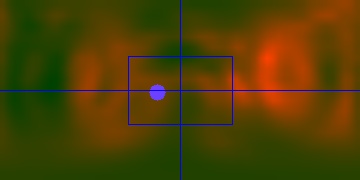}%
\includegraphics[width=0.166\linewidth]{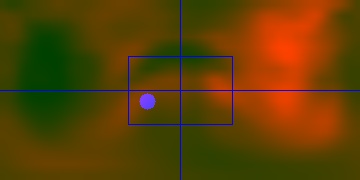}%
\includegraphics[width=0.166\linewidth]{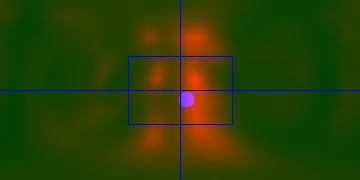}%
\includegraphics[width=0.166\linewidth]{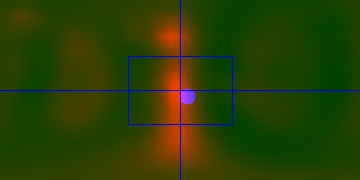}%
\linebreak
\includegraphics[width=0.166\linewidth]{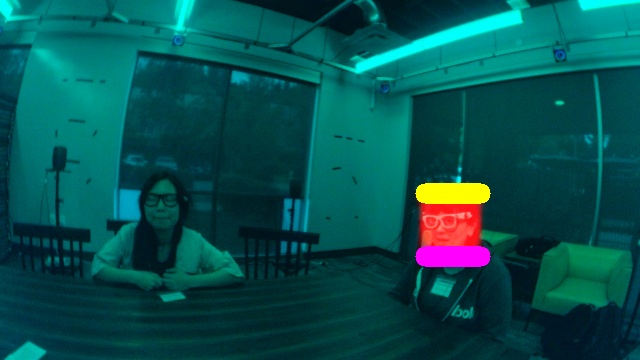}%
\includegraphics[width=0.166\linewidth]{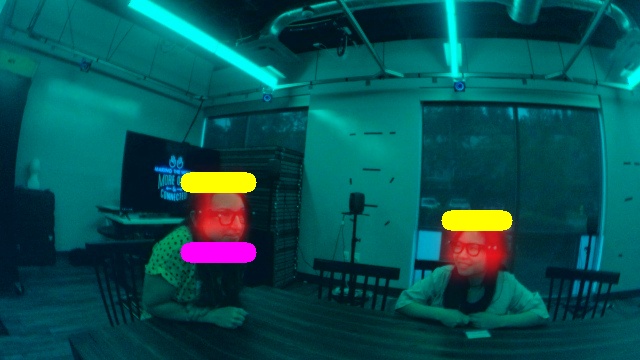}%
\includegraphics[width=0.166\linewidth]{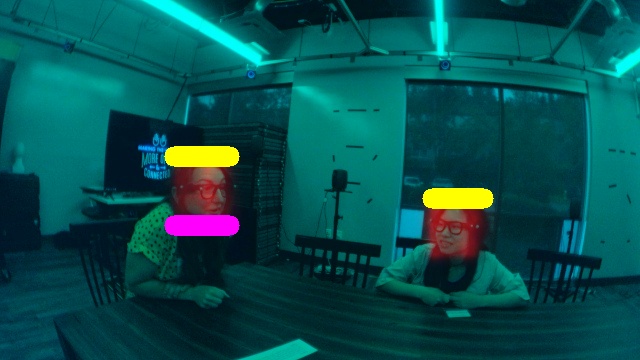}%
\includegraphics[width=0.166\linewidth]{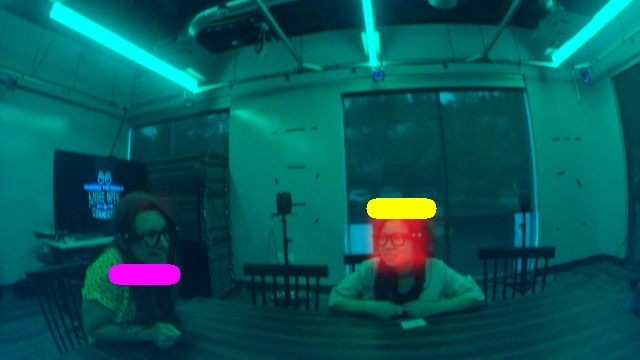}%
\includegraphics[width=0.166\linewidth]{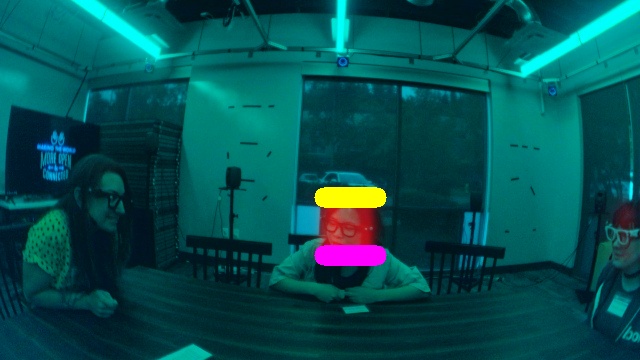}%
\includegraphics[width=0.166\linewidth]{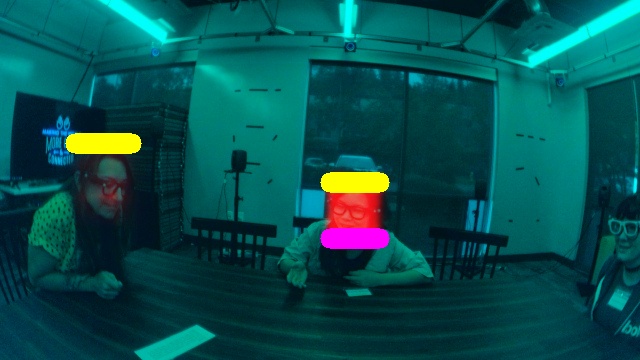}%
\linebreak
\includegraphics[width=0.166\linewidth]{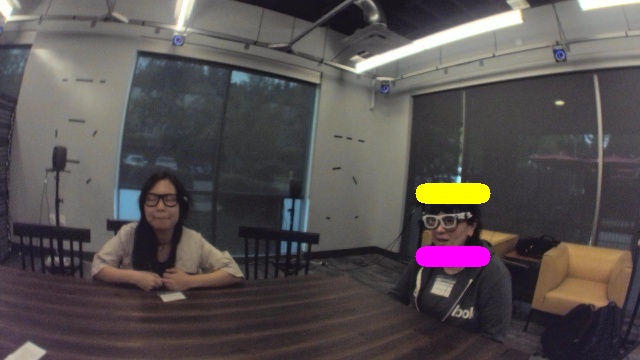}%
\includegraphics[width=0.166\linewidth]{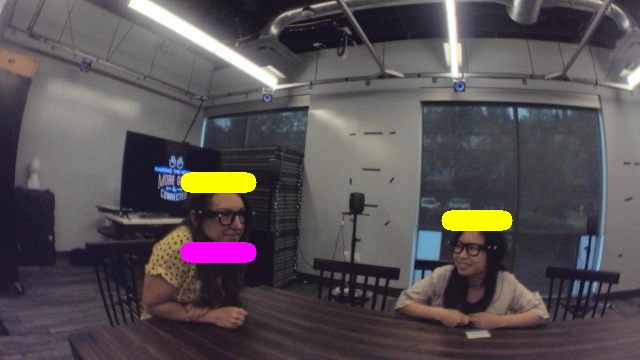}%
\includegraphics[width=0.166\linewidth]{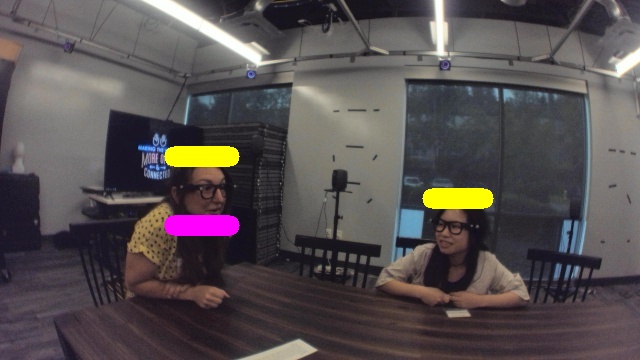}%
\includegraphics[width=0.166\linewidth]{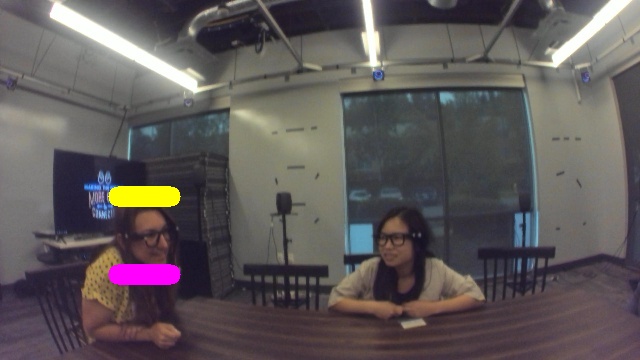}%
\includegraphics[width=0.166\linewidth]{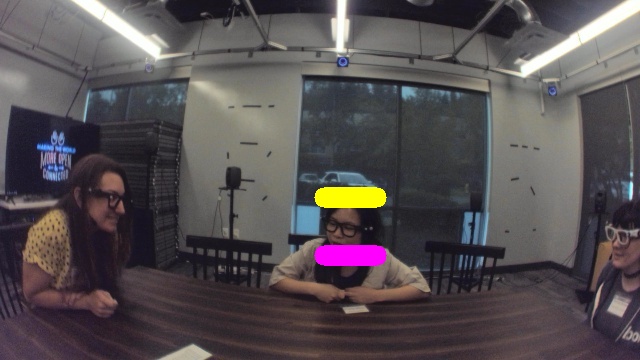}%
\includegraphics[width=0.166\linewidth]{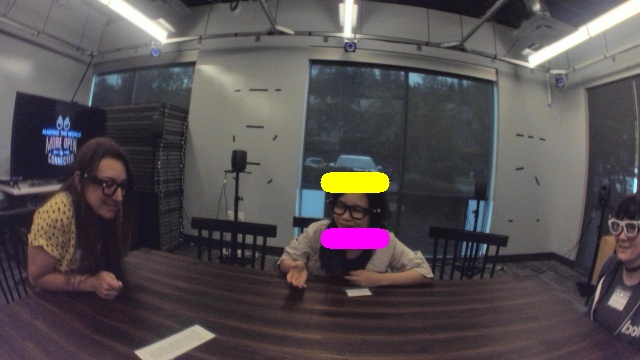}%
\linebreak
\includegraphics[width=0.166\linewidth]{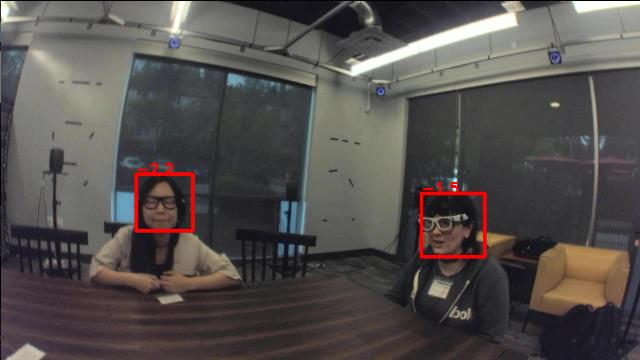}%
\includegraphics[width=0.166\linewidth]{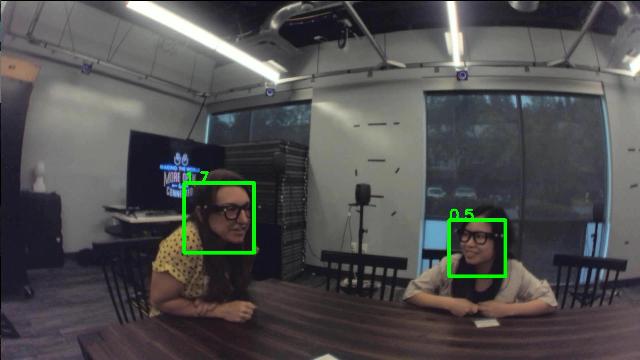}%
\includegraphics[width=0.166\linewidth]{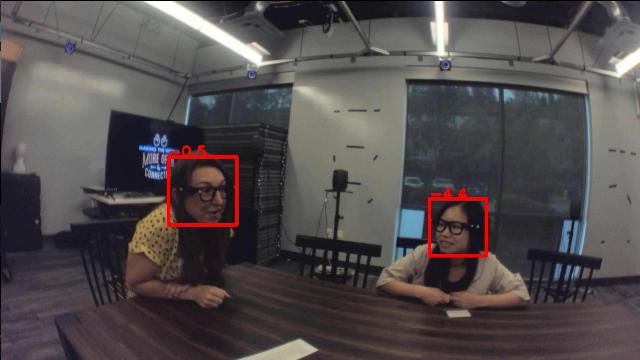}%
\includegraphics[width=0.166\linewidth]{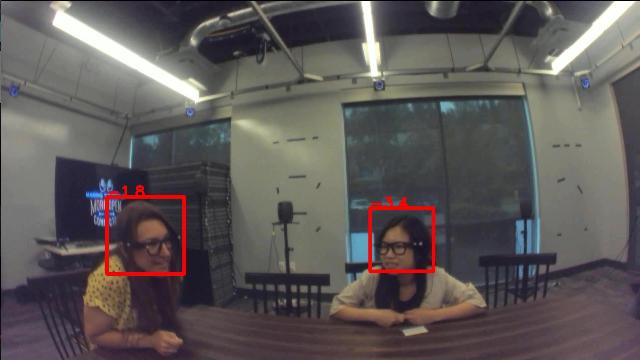}%
\includegraphics[width=0.166\linewidth]{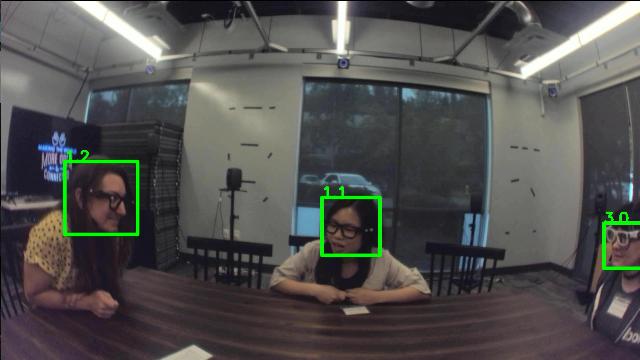}%
\includegraphics[width=0.166\linewidth]{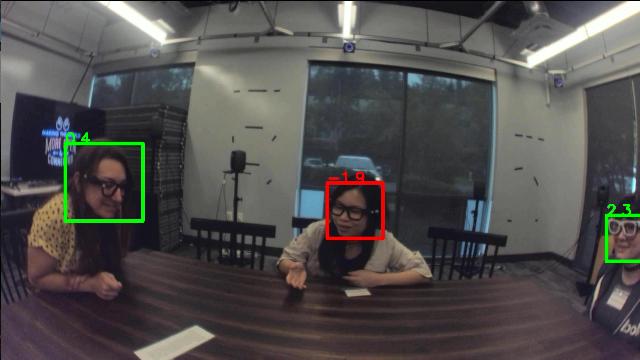}%
\vspace{-5pt}
	\caption{ Qualitative comparison results. The purple bar indicates when a person is predicted
	to be talking while the yellow bar is the corresponding ground truth.
	Rows 2, 4: the predicted 360$^{\circ}$ voice map compared against the the ground truth in blue channel. 
	Rows 1, 2: The result of \texttt{Ours AV(corr)}. Rows 3, 4: \texttt{DOA+headbox}, Row 5: \texttt{DOA+image},   
	Row 6: \texttt{MRC(EasyCom)}, Row 7: \texttt{TalkNet(EasyCom)}. In Row 7, green boxes indicate active speech
	while red boxes are inactive.
	}
\vspace{-10pt}	
\label{fig:comparison}	
\end{figure*}

We first evaluate the mean average precision (mAP)
of active speaker localization detections
within the camera's field of view. We compare against multi-channel as well as one- and two-channel 
audio-visual methods and visual-only method.
The mAP is computed based on the scores
within the 
ground truth head bounding boxes in each video frame.
For our methods and the competing methods \texttt{DOA+headbox}, \texttt{DOA+image}, \texttt{AV-rawaudio}, and \texttt{BinauralAVLocation} 
we extract the voice heat map's maximum value in each ground truth head bounding box and use it 
as the detection score. The \texttt{MRC} and 
the \texttt{TalkNet} methods use the classification probability of the corresponding head box as the detection score. 
Both \texttt{MRC} and \texttt{TalkNet} use the ground truth head bounding boxes for testing. 

As shown in Table ~\ref{tab:comp1}, our methods give 
much higher mAP than all of the competing methods.
Fig.~\ref{fig:comparison} shows qualitative comparison results.
Due to the difficulty in learning useful features from raw audio, \texttt{AV-rawaudio} gives inferior results
in comparison to spectrogram and cross-correlation audio features. 
Background noise also causes  
traditional audio-only signal processing approaches give blurry DOA maps and 
inaccurate target localization results. 
The \texttt{DOA+image} deep learning method that combines this DOA map with video frames improves performance, but
  still gives lower mAP than our proposed method. 
	This emphasizes the benefit of learning spatial audio-visual representations end-to-end.
Our method also gives much higher mAP than the previous video-only \texttt{MRC} and single-channel audio-visual active speaker detection method \texttt{TalkNet}
trained on both the AVA dataset \cite{avadataset} and the EasyCom dataset. 
Our method greatly outperforms the \texttt{BinauralAVLocation} in both the 4-channel and 2-channel audio settings.

\setlength{\tabcolsep}{1pt}
\begin{table}[tb]	
  \small
  \centering	
  \begin{tabular}{ c | c  }
    \hline
			       & ASL mAP \\ \hline
	   Ours AV(cor)         & 84.14      \\
	   Ours AV(cor+eng)     & 83.32      \\
           Ours AV(cor+box)      & 86.25       \\	   
	   Ours AV(cor+eng+box)  & 86.32       \\
	   Ours AV(spec)  & 85.49      \\
	   Ours AV (eng)         & 62.68       \\	   
	   Ours AV(cor)-2ch      & 80.00       \\
	   Ours AV(spec)-2ch     & 83.30               \\
	   \hline
	   AV-rawaudio        & 72.32       \\
	   DOA+headbox             & 52.62      \\
	   DOA+image           & 54.27      \\ 
	   MRC (AVA)           & 46.60      \\ 
	   MRC(EasyCom)        & 64.24      \\ 
	   TalkNet (AVA)       & 69.13      \\ 
	   TalkNet (EasyCom)   & 44.24      \\
	   BinauralAVLoc         & 60.75      \\
    \hline
  \end{tabular}
	\caption{Comparison of mAPs in the visual field of view. Most of these tests use 4-channel audio, except
	for \texttt{Ours AV(cor)-2ch}, \texttt{Ours AV(spec)-2ch}, \texttt{BinauralAVLoc}, which use 2-channel audio, \texttt{TalkNet} which uses
	single-channel audio, and video-only \texttt{MRC}.
	}
	\label{tab:comp1}
\vspace{-10pt}	
\end{table}

For different variations of the proposed method, as shown in Table~\ref{tab:comp1}, the energy feature is significantly worse than the other two features, 
while spectrogram features give slightly better mAP. 
The cross correlation and energy features are still attractive due to their speaker-invariant 
properties and thus have potential to generalize better in real applications and preserve privacy. 
The cross correlation feature is also invariant to the microphone gain settings;
this makes it useful when the gains need to change dynamically for best signal-noise ratio.

We also compare our audio-only and video-only variations with the full audio-visual model.
In comparison to our full audio-visual method \texttt{Ours AV(cor+mag+box)} with a mAP of 86.32\%, 
the video-only variation gave a much lower mAP of 58.44\% and the 
audio-only version also gave a lower mAP of 78.08\%. 
The results of \texttt{Ours AV(corr+box)} and  \texttt{Ours AV(corr+eng+box)} also show that our proposed method
can generalize to different environments by removing background visual information outside of head detections,
which can potentially 
improve the result.
Even with only two audio channels, our network still gave strong results that outperformed the \texttt{BinauralAVLoc}
network architecture designed to leverage the symmetry of binaural audio.

\subsection{Spherical Active Speaker Localization} 

One unique property of our proposed method is that it gives a full 360$^{\circ}$ spherical speaker localization result. 
Since there is no head bounding box outside of the field of view, we use the angular error to 
measure the localization quality. 

The metric is defined as follows:
We first extract the detected target locations in the predicted 
voice heat map using non-maximum suppression. Every peak in the heat map with value greater than a threshold is a potential target. 
In the experiments, we set the threshold to 0.
The positions in the heat map indicate the angles of directions.
We compute the minimum distances from the detected points to the ground truth points in the voice heat map, 
whose mean is denoted as E1. We compute mean E1 and its standard deviation
Std1. The corresponding metrics from the ground truth point set to the detected point set are mean E2 and Std2. 
The reason we use distance metric in two directions is to take both missing detections and false alarms into account.

Not all the competing methods can give full 360$^{\circ}$ spherical localization results.
In this experiment, we compare our method with the methods that use traditional DOA maps and the audio-visual variation with raw audio input.
As shown in Table~\ref{tab:vloc360}, our method gives the lowest angular errors. 

\setlength{\tabcolsep}{1pt}
\begin{table}[tbh]
  \small	
  \centering
  \begin{tabular}{ c | c | c | c | c }
    \hline
				     & Mean E1 & Std1   & Mean E2 & Std2  \\ \hline
                      Ours AV (cor) & 16.77   & 12.63  & 6.56   & 8.77 \\
		      Ours AV (spec) & 8.81   & 9.63  & 6.21   & 6.89 \\
		      DOA            & 129.82  & 18.26  & 46.45   & 21.50 \\
                      DOA+image       & 66.81  & 7.89  & 36.48   & 8.97 \\
		      AV-rawaudio    & 40.14     &  10.55    &  140.75     &  19.58   \\
    \hline
  \end{tabular}
  \caption{Comparison of full 360$^{\circ}$ spherical voice activity localization errors measured in degrees.}
        \label{tab:vloc360}
\vspace{-10pt}	
\end{table}

\setlength{\tabcolsep}{1pt}
\begin{table}[tbh]	
  \small	
  \centering	
  \begin{tabular}{ c | c  }
    \hline
		      & Wearer Audio activity mAP  \\ \hline
		      Ours(cor) & 90.20  \\               
	              Ours(cor+eng) & 90.13  \\    
		      Ours(eng) & 88.89   \\
		      Ours(spec) & 91.69  \\
		      Ours(cor)-2ch & 87.66   \\
		      Ours(spec)-2ch & 90.14     \\
		      Eng(single channel) & 76.71 \\
		      AV-rawaudio & 87.29 \\

    \hline
  \end{tabular}
	\caption{Camera wearer voice activity detection. \texttt{Eng(single channel)} is the 
	naive approach of using short-time energy for wearer voice classification.}
	\label{tab:wearer}
\vspace{-10pt}	
\end{table}

\vspace{-5pt}
\subsection{Wearer Speech Activity Detection}
\vspace{-5pt}
Another unique property of the proposed method is that it can simultaneously detect the voice activity of the person wearing the recording glasses. 
Our method shares the learned audio 
features for both tasks. 
During the training of the camera wearer voice networks, 
the shared feature design freezes the network feature extraction parameters while only training the last two fully connected layers.

Camera wearer audio activity detection is a new task. We construct different natural solutions in the comparison.
Table~\ref{tab:wearer} summarizes the comparison result.
As shown in Table~\ref{tab:wearer}, our proposed method gives better results than the competing methods.
The shared feature design in fact also gives better result than training a separate wearer voice classification model.
For instance, our method using cross correlation input features gives 90.2\% mAP, but if we retrain a separate wearer classifier 
the mAP is 88.01\%.
This is likely because of the additional supervision in training the localization task to explicitly suppress the wearer's speech. 

Comparing to traditional signal processing approaches, our method requires more computationally expensive GPU operations.
However, the proposed method is still efficient. It runs in real time at over 180 frames per second 
using a single GTX2080Ti GPU with about 50\% utilization. 
More optimization could also further improve
the efficiency of the network. The proposed method also has a smaller latency compared to traditional signal processing methods, 
which require estimating signal statistic over longer windows of time.
While we only use 4 microphones in our experiments, the proposed method could be easily extended to devices with any number of microphones 
in any array configuration. With a larger microphone array, the proposed method has the potential to achieve even better results. 

\vspace{-5pt}
\section{Conclusion}
\vspace{-5pt}
We proposed a novel multi-channel audio-visual method to tackle the 360$^{\circ}$ spherical active speaker detection problem
for localizing active speakers both within and beyond an egocentric camera's visual field of view while also simultaneously predicting
the wearer's voice activity. 
Our experiments showed that the proposed method gives superior results to competing methods
and can run in real time with short latency. It can be deployed to enable many useful AR functions. 



\small
\begin{thebibliography}{9}
\bibitem{360sound} P. Morgado, N. Vasconcelos, T. Langlois, O. Wang.
Self-Supervised Generation of Spatial Audio for 360-degree Video,
NIPS 2018.

\bibitem{wangaaai} X. Wu, Z. Wu, L. Ju, S. Wang. 
Binaural Audio-Visual Localization,
AAAI-21.

\bibitem{av1} A. Owens,     A. A. Efros   
Audio-Visual Scene Analysis with Self-Supervised Multisensory Features,
ECCV 2018.

\bibitem{avsep1} A. Senocak, T.-H. Oh, J. Kim, M.-H. Yang, I. S. Kweon. Learning to Localize Sound Source in Visual Scenes, CVPR 2018.

\bibitem{avsep2} A. Ephrat, I. Mosseri, O. Lang, T. Dekel, K. Wilson, A. Hassidim, W. T. Freeman, M. Rubinstein. 
	Looking to Listen at the Cocktail Party: A Speaker-Independent Audio-Visual Model for Speech Separation,
ACM Transactions on Graphics, Vol. 37, No. 4, pp 1-11, August 2018.

\bibitem{avsep4} R. Gao, R. Feris, K. Grauman. Learning to Separate Object Sounds by Watching Unlabeled Video, CVPR 2018.

\bibitem{avsep5} T. Afouras, J.S. Chung, A. Zisserman. The Conversation: Deep Audio-Visual Speech Enhancement,  arXiv:1804.04121. 

\bibitem{avloc1}
I. D. Gebru, X. Alameda-Pineda, R. Horaud, F. Forbes.
Audio-visual speaker localization via weighted clustering,
IEEE International Workshop on Machine Learning for Signal Processing (MLSP), 2014.

\bibitem{avloc2}
R. Qian, D. Hu, H. Dinkel, M. Wu, N. Xu, and W. Lin.
Multiple Sound Sources Localization from Coarse to Fine,
ECCV 2020.

\bibitem{avloc3}
H. Chen, W. Xie, T. Afouras, A. Nagrani, A. Vedaldi, A. Zisserman. 
Localizing Visual Sounds the Hard Way,
CVPR 2021.

\bibitem{soundloc1}
C. Rascon, I. Meza. Localization of Sound Sources in Robotics: A Review.
Robotics and Autonomous Systems, Volume 96, October 2017, Pages 184-210.

\bibitem{deep-sound-loc1}
S. Adavanne, A. Politis, T. Virtanen.
Direction of Arrival Estimation for Multiple Sound Sources Using Convolutional Recurrent Neural Network, European Signal Processing Conference (EUSIPCO), 2018.

\bibitem{deep-sound-loc2}
T.N.T. Nguyen, W-S. Gan, R. Ranjan, D.L. Jones.
Robust Source Counting and DOA Estimation Using Spatial Pseudo-Spectrum and Convolutional Neural Network, IEEE/ACM Transactions on Audio, Speech, and Language Processing ( Volume: 28)

\bibitem{talknet} R. Tao, Z. Pan, R.K. Das, X. Qian, M.Z. Shou, and H. Li. Is Someone Speaking? Exploring Long-term Temporal Features for Audio-visual Active Speaker Detection. The 29th ACM International Conference on Multimedia, 2021.

\bibitem{iccv21a} O. Kopuklu, M. Taseska, G. Rigoll.   	
How To Design a Three-Stage Architecture for Audio-Visual Active Speaker Detection in the Wild,
ICCV 2021.

\bibitem{iccv21b} T.-D. Truong,
C. N. Duong,
T. D. Vu,
H. A. Pham,
B. Raj,
N. Le
K. Luu.
The Right to Talk:
An Audio-Visual Transformer Approach. ICCV 2021.

\bibitem{avsep6} R. Gao and K. Grauman. VisualVoice: Audio-Visual Speech Separation with Cross-Modal Consistency. 
	CVPR 2021.

\bibitem{easycom} J. Donley, V. Tourbabin, J. Lee, M. Broyles, H. Jiang, J. Shen, M. Pantic, V. K. Ithapu, R. Mehra.
EasyCom: An Augmented Reality Dataset to Support Algorithms for Easy Communication in Noisy Environments,
	arXiv:2107.04174
	
\bibitem{deepdoasurvey} P.-A. Grumiaux, S. Kitic, L. Girin, A. Guerin.	
A Survey of Sound Source Localization with Deep Learning Methods,
arXiv:2109.03465.

\bibitem{doa} V. Tourbabin, J. Donley, B. Rafaely, R. Mehra. Direction of Arrival Estimation in Highly Reveberant Environments
Using Soft Time-Frequency Mask, 2019 IEEE Workshop on Applications of Signal Processing to Audio and Acoustics.

\bibitem{audioprocess} D. P. Jarrett, E. A.P. Habets, P. A. Naylor. Theory and Applications of Spherical Microphone Array Processing,
Springer Topics in Signal Processing, 9.

\bibitem{avadataset} J. Roth, S. Chaudhuri, O. Klejch, R. Marvin, A. Gallagher, L. Kaver, S. Ramaswamy, A. Stopczynski, C. Schmid, Z. Xi, C. Pantofaru. 
	AVA-ActiveSpeaker: An Audio-Visual Dataset for Active Speaker Detection, arXiv:1901.01342.

\bibitem{voxconverse} J. S. Chung, J. Huh, A. Nagrani, T. Afouras, A. Zisserman. Spot The Conversation: Speaker Diarisation in The Wild, ArXiv, 2020.	

\bibitem{voxceleb2} J. S. Chung, A. Nagrani, A. Zisserman. VoxCeleb2: Deep Speaker Recognition, INTERSPEECH, 2018.

\bibitem{avspeech} T. Afouras, 
J. S. Chung, 
A. Senior, 
O. Vinyals, 
A. Zisserman. 
Deep Audio-visual Speech Recognition,
TPAMI, December, 2018.

\bibitem{avnavi} C. Chen, U. Jain, C. Schissler, S. V. Amengual Gari, 
Z. Al-Halah, V. K. Ithapu, P. Robinson, K. Grauman.
SoundSpaces: Audio-Visual Navigation in 3D Environments,
ECCV 2020.

\bibitem{kazakos2019epic} E. Kazakos, A. Nagrani, A. Zisserman, D. Damen. 
	EPIC-Fusion: Audio-Visual Temporal Binding for Egocentric Action Recognition.
	ICCV 2019.

\bibitem{audiovisual-slowfast} F. Xiao, Y. J. Lee, K. Grauman, J. Malik, C. Feichtenhofer.
	Audiovisual SlowFast Networks for Video Recognition, arXiv, 2020.
	
\end{thebibliography}
\end{document}